# Generalizable 7T T1-map Synthesis from 1.5T and 3T T1 MRI with an Efficient Transformer Model

Zach Eidex[1], Mojtaba Safari[1,2], Tonghe Wang[3], Vanessa Wildman[1], David S. Yu[1,5], Hui Mao[4,5], Erik Middlebrooks[6], Aparna Kesarwala[1,5*] and Xiaofeng Yang[1,2,5*]

[1]Department of Radiation Oncology, Emory University, Atlanta, GA

[2]Department of Computer Science and Informatics, Emory University, Atlanta, GA

[3]Department of Medical Physics, Memorial Sloan Kettering Cancer Center, New York, NY

[4]Department of Radiology and Imaging Sciences, Emory University, Atlanta, GA

[5]Winship Cancer Institute, Emory University, Atlanta, GA

[6]Department of Radiology, Mayo Clinic, Jacksonville, FL

*Email: xiaofeng.yang@emory.edu and aparna.kesarwala@emory.edu

## ABSTRACT

**Background:** Ultra-high-field 7 Tesla (7T) magnetic resonance imaging (MRI) provides improved resolution and contrast over standard clinical field strengths (1.5T, 3T). However, 7T scanners are costly, scarce, and introduce additional challenges such as susceptibility artifacts.

**Purpose:** We propose an efficient transformer-based model (7T-Restormer) to synthesize 7T-quality MRI (quantitative T1MP2RAGE maps) from routine 1.5T or 3T T1-weighted (T1W) images.

**Methods:** The proposed method leverages an efficient restoration transformer backbone and spatial attention layers to capture long-range dependencies and generate high-quality 7T-like images. Our model was validated on an institutional dataset, comprised of 35 1.5T and 108 3T T1w MRI paired with corresponding 7T T1 maps of patients with confirmed multiple sclerosis (MS). A total of 141 patient cases (32,128 slices) were randomly divided into 105 (25; 80) training cases (19,204 slices), 19 (5; 14) validation cases (3,476 slices), and 17 (5; 14) test cases (3,145 slices) where (X; Y) denotes the patients with 1.5T and 3T T1W scans, respectively. The synthetic 7T T1 maps were evaluated by comparing their similarity to the ground truth volumes with the following metrics: peak signal-to-noise ratio (PSNR), structural similarity index measure (SSIM), and normalized mean squared error (NMSE) and compared against the ResViT and ResShift models. Statistical significance ($p < .05$) was determined with the Welch's t-test.

**Results:** The 7T-Restormer model outperforms state-of-the-art approaches while using fewer parameters, achieving a PSNR of 26.0 ± 4.6 dB, SSIM of 0.861 ± 0.072, and NMSE of 0.019 ± 0.011 for 1.5T inputs, and 25.9 ± 4.9 dB, and 0.866 ± 0.077 for 3T inputs, respectively. Using 10.5 M parameters, our model reduced NMSE by 64 % relative to 56.7M parameter ResShift (0.019 vs 0.052, $p = <.001$ and by 41 % relative to 70.4M parameter ResViT (0.019 vs 0.032, $p = <.001$) at 1.5T, with similar advantages at 3T (0.021 vs 0.060 and 0.033; $p < .001$). Training with a mixed 1.5 T + 3 T corpus was superior to single-field strategies. Restricting the model to 1.5T increased the 1.5T NMSE from 0.019 to 0.021 ($p = 1.1 \times 10^{-3}$) while training solely on 3T resulted in lower performance on input 1.5T T1W MRI.

**Conclusion:** We propose a novel method for predicting quantitative 7T MP2RAGE maps from 1.5T and 3T T1W scans with higher quality than existing state-of-the-art methods. Future development and application of our approach may enhance diagnostic accuracy, treatment planning, and facilitate downstream tasks by making the benefits of 7T MRI more accessible to standard clinical workflows.

## 1. INTRODUCTION

Magnetic resonance imaging (MRI) at the ultra-high field, i.e., 7 Tesla (7T), offers enhanced spatial resolution compared to conventional MRI at lower field strengths, enabling superior visualization of anatomical structures and pathological features.[1] Despite these advantages, the availability and clinical adoption of 7T MRI is limited due to high costs, demand on technical expertise, concern over the safety for patients with metal implants, and vulnerability to imaging artifacts such as field (B0) and transmitter (B1) inhomogeneities and motion artifacts.[2] Thus, there is great interest and increasing effort in leveraging widely available images obtained from current clinical MRI systems at 1.5 or 3T to generate high resolution and noise-free synthetic 7T-like images using deep learning approaches. However, early studies using convolutional neural networks (CNNs)[8] struggle to capture long-range context due to their local receptive fields. While vision transformer (ViT) architectures address this limitation by enabling global context through self-attention mechanisms,[9-11] self-attention introduces quadratic, $O(n^2)$, computational complexity often making ViT computationally prohibitive in medical imaging. Hybrid CNN-transformer architectures[12-14] and Swin transformers[15,16] reduce computational burden at the expense of limiting the receptive field in shallower layers. Other generative approaches, including generative adversarial networks (GANs)[17-19] and diffusion models[20-22], suffer from training instability and can have inconsistent performance while the multiple timesteps in diffusion models lead to lengthy inference times.

We propose the 7T-Restormer model to synthesize 7T quality images, even quantitative relaxometry maps, directly from images obtained from 1.5T and 3T MRI scanners. In this study, we are the first to use T1-weighted images from 1.5 and 3T to generate the spatial mapping of tissue longitudinal relaxation times, or T1 maps, matching those obtained at 7T using the acquisition pulse sequence of Magnetization Prepared 2 Rapid Acquisition Gradient Echoes (MP2RAGE).{Marques, 2010 #48} MP2RAGE improves T1 mapping by reducing the sequence sensitivity to transmitter and field inhomogeneities[3] at 7T, making it particularly beneficial for accurate cortical parcellation[4] to improve characterization of neurological diseases like multiple sclerosis (MS)[5,6], and quantitative monitoring of neurodegenerative conditions.[7] Our model is based on a Restormer backbone which maintains a global receptive field by applying self-attention in the channel direction as opposed to across the spatial dimensions, reducing the complexity from $O(n^2)$ to $O(n)$.[23] To evaluate the performance of our architecture we compare our results to the state-of-the-art GAN-based ResViT[24] and diffusion-based ResShift architectures.[25] In addition, we evaluate the impact of training our model on a mixed-field-strength dataset compared to single-field training and find that training 1.5T and 3T MRI together results in better performance than training on each field strength individually. Our lightweight transformer-based model achieves state-of-the-art performance in predicting 7T T1-maps while mitigating the training instability and slow inference that characterize existing GAN- and diffusion-based methods.

## 2. METHODS

### 2.1 Problem Formulation

We approach ultra-high-field synthesis as a supervised image-to-image translation problem. Given an input axial slice $x \in \mathbb{R}^{H \times W}$ from a routine low-field scan (either 1.5T or 3T, T1-weighted MRI) and its paired 7T quantitative T1-map $y$, we aim to learn a function $F_\theta : x \mapsto \hat{y}$ that produces an output $\hat{y}$ minimizing the voxel-wise $L_1$ loss $|\hat{y} - y|_1$. The $L_1$ loss was empirically found to yield lower normalized mean squared error (NMSE) compared to combined adversarial and L1 losses. In addition, the risk of generating hallucinated textures was reduced and training stability was improved at the expense of slight image smoothing in regions of uncertainty.[26]

### 2.2 Data Acquisition and Preprocessing

This retrospective study was approved by the IRB of Mayo Clinic. The dataset used in this study consisted of paired scans from 141 patients with a confirmed diagnosis of MS.[27] 108 patients had both 3T and 7T scans, while 35 patients had 1.5T and 7T scans. Due to severe artifacts, 2 patients were removed from the test dataset. The T1 maps were acquired on a 7T scanner (Siemens MAGNETOM Terra) with a 8-channel transmit/32-channel receive head coil using a 3D Low Angle Minimizing Artifacts MP2RAGE sequence with the following key imaging parameters: TR = 4.5 s, TE = 2.2 ms, TI1/TI2 = 0.95/2.5 s, FA1/FA2 = 6∘/4∘, FOV = 230 × 230 $cm^2$, matrix size of 288 × 288, a resolution of 0.8 × 0.8 × 0.8 mm3, in the sagittal view. The 1.5T and 3T T1W MRI were from various sequences and scanners which may improve the generalizability of the model. Preprocessing steps included rigidly aligned the lower field MRI to the 7T image space and resampled to 0.8 mm isotropic resolution. Skull stripping was performed by extracting the brain masks with FSL BET.[28]. 2D axial slices from the 3D volumes were center-cropped to a of size $256 \times 384$ and normalized to the range [-1,1] to accommodate the network architectures. Finally, a Lipari color map was applied to the 7T images as proposed in Furderer et al.[29]

## 2.3 Model Architecture

The proposed 7T-Restormer model leverages the Restormer architecture, reducing computational complexity from quadratic ($O(n^2)$) to linear ($O(n)$) by first embedding depthwise convolutions into the query, key, and value projections before computing self-attention. This shifts the self-attention mechanism from spatial to channel dimensions, enabling global context and allowing self-attention to be directly applied to large medical images without limiting the receptive field.[23,30]

### 2.3.1 Encoder and Decoder

Our architecture employs an encoder-decoder framework with spatial attention computed at the bottleneck. The first layer of the encoder is a 3×3 overlapping convolution, transforming the single-channel input image into a 48-dimensional feature representation. This is followed by a Restormer block to refine features. Subsequently, the feature maps are downsampled via a pixel-unshuffle operation[31], halving the spatial dimensions and expanding the channels to 96. This process is repeated, reducing spatial dimensions by another factor of two and increasing channels to 192. Each encoder level incorporates one or two Restormer blocks, culminating in an additional global MDTA block at the bottleneck that thoroughly captures spatial context through spatial attention mechanisms (Figure 1). In the decoder, spatial resolution is sequentially restored using pixel-shuffle operations, which double the spatial dimensions at each level. At each hierarchical decoder stage, encoder features are concatenated with decoder features via skip connections, and Restormer blocks further refine these combined representations. Finally, a convolutional layer followed by a hyperbolic tangent (tanh) activation function outputs synthesized 7T T1-maps, normalized within the intensity range [-1, 1].

### 2.3.2 Multi-Dconv Head Transposed Attention (MDTA)

The MDTA block modifies the traditional self-attention mechanism by embedding depthwise convolutions into query, key, and value projections. The features first pass through a normalization layer. Each projection then undergoes a 1×1 point-wise convolution to integrate cross-channel information, followed by a 3×3 depthwise convolution to encode spatially local context. Attention is subsequently computed across the channel dimension, significantly reducing computational load. A softmax function is applied to generate an attention map, and the resulting features are processed through a 1×1 convolutional layer and combined with the original input with a skip connection.

2.3.3 Gated-Dconv Feed-Forward Network (GDFN)

While MDTA primarily captures global context, the GDFN block further refines local details. The GDFN employs a gating mechanism consisting of a normalization layer followed by two parallel pathways with depthwise separable convolutions. One pathway uses a Gaussian error linear unit (GELU) activation to selectively activate relevant features, while the other serves as a linear pass-through.[32] Outputs from both pathways are combined through element-wise multiplication and pass through a 1×1 convolutional layer before being connection to the original feature maps via skip connection.

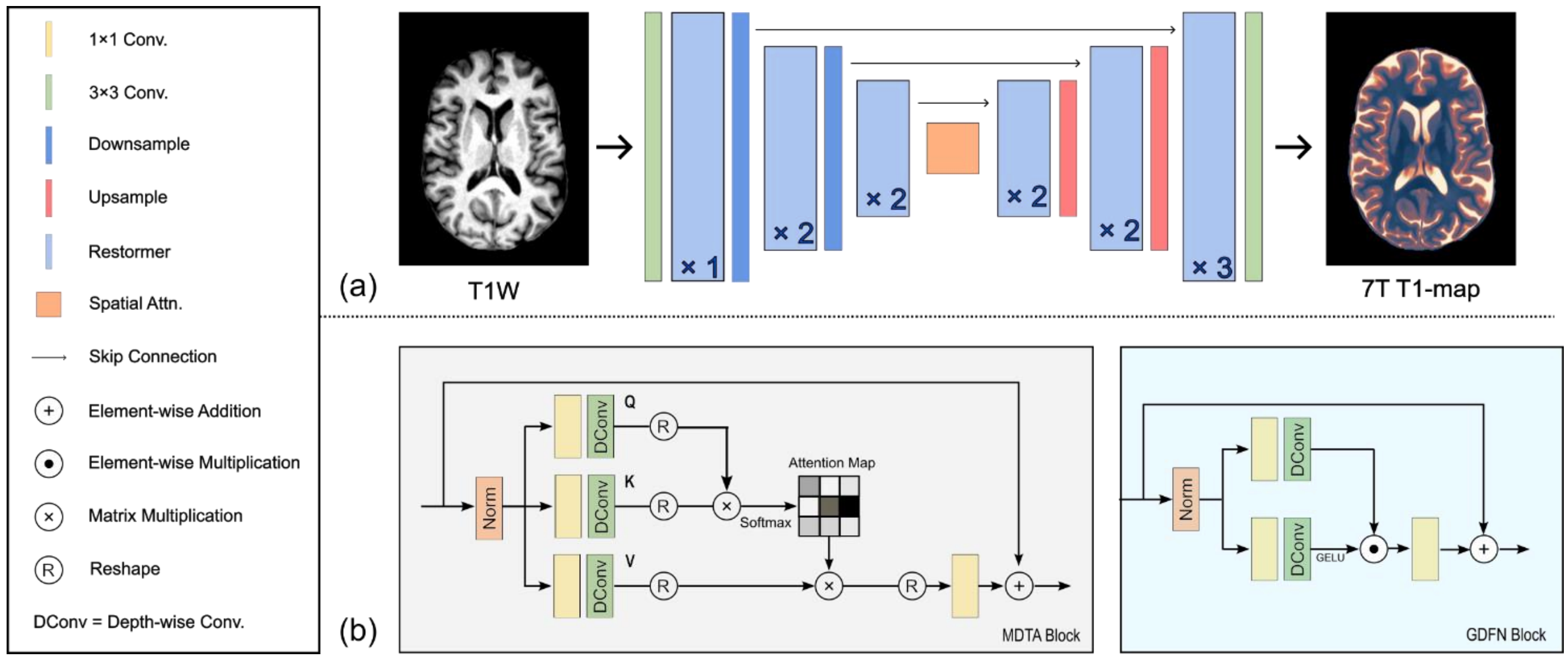

**Figure 1**. Schematic flow chart of the 7T-Restormer model. (a) The model processes input 1.5T or 3T T1-weighted MRI slices through hierarchical encoder-decoder stages incorporating Restormer blocks. "× N" denotes multiple Restormer blocks connected in sequence. The downsampling and upsampling blocks are comprised of convolutional and pixel shuffle/unshuffle layers. (b) Each Restormer block utilizes sequential MDTA and GDFN modules. Notably, the depth-wise convolutions (DConv) shift self-attention from the spatial dimensions to the channel dimension.

2.4 Model Implementation

The 7T-Resformer model was trained from scratch with the AdamW optimiser (learning rate $1 \times 10^{-4}$, $\beta_1 = 0.9$, $\beta_2 = 0.999$, weight-decay $1 \times 10^{-4}$.[33] The model was trained at full precision for 50 epochs with a batch size of 8 on a single NVIDIA A6000 ADA (48 GB). During training, the data was augmented by randomly flipping the axial slices across the axial plane. The competing methods (ResViT[24] and ResShift[25]) were trained according to the hyperparameters and learning schedule as proposed in the original work. For ResViT, the transformer layers were pretrained for 100 epochs followed by 50 epochs of finetuning the model. ResShift was trained for 100 epochs.

2.5 Evaluation and Validation

A hold-out test was performed to assess model performance. The full cohort of 141 paired cases was randomly split into 105 (25; 80) training cases (19,204 slices), 19 (5; 14) validation cases (3,476 slices), and 17 (5; 14) test cases (3,145 slices) where (X; Y) denotes the patients with 1.5T and 3T T1W scans, respectively. All quantitative results reported below are computed on the slice-wise predictions from the 19-patient test set. Quantitative performance was calculated with normalized mean-squared error (NMSE), peak signal-to-noise ratio (PSNR), and structural similarity index measure (SSIM). Statistical significance of the results was evaluated with a two-tailed Welch's t-test to evaluate whether the proposed 7 T-Resformer outperformed baseline methods (ResShift and ResViT).[34] The null hypothesis stated that the mean metric value of Resformer equals that of the baseline; the alternative hypothesis stated that they differ. A significance level of α=0.05 was adopted, therefore *p*-values below 0.05 were deemed to be statistically significant.

### 2.5.1 Image-quality metrics

Synthetic 7T T1-maps, $\hat{y}$, were compared with ground-truth 7T references, y, using three whole-volume measures - NMSE, PSNR, and SSIM as defined in Eqs. (1)–(3). NMSE quantifies the voxel-wise residual energy after normalization by the energy of the reference volume.[35] A value of zero corresponds to perfect reconstruction and increases monotonically with error. PSNR expresses the mean-squared error on a logarithmic decibel scale, yielding a similarity score that aligns more closely with perceived image fidelity.[36] A higher PSNR indicates greater correspondence between synthetic and reference images. SSIM evaluates luminance, contrast, and structural congruence to approximate human visual perception; it ranges from –1 to 1 and attains 1 when two images are identical.[37] Before metric computation, all intensities were linearly rescaled to [−1,1], giving a dynamic range of L=2. The stabilizing constants in Eq. (3) were set to $C_1 = (0.01L)^2$ and $C_2 = (0.03L)^2$.[37] Under this convention, larger PSNR and SSIM values and smaller NMSE values correspond to superior synthesis accuracy.[38]

$$\mathrm{NMSE} = \frac{\sum_{i=1}^{N}(y_i - \hat{y}_i)^2}{\sum_{i=1}^{N} y_i^2} \tag{1}$$

$$\mathrm{PSNR} = 10\,log\left(\frac{L^2}{\frac{1}{N}\sum_{i=1}^{N}(y_i - \hat{y}_i)^2}\right) \tag{2}$$

$$\mathrm{SSIM}(y, \hat{y}) = \frac{(2\mu_y\mu_{\hat{y}} + C_1)(2\sigma_{y\hat{y}} + C_2)}{(\mu_y^2 + \mu_{\hat{y}}^2 + C_1)(\sigma_y^2 + \sigma_{\hat{y}}^2 + C_2)} \tag{3}$$

## 3. RESULTS

Quantitative results are summarized in Table 1. For 1.5T inputs, our proposed 7T-Restormer method achieved an NMSE of 0.019 ± 0.011 (mean ± SD), PSNR of 25.98 ± 4.57 dB, and SSIM of 0.861 ± 0.072. For 3T inputs, the corresponding values were 0.021 ± 0.015, 25.89 ± 4.86 dB, and 0.866 ± 0.077. Across both field strengths 7T-Restormer provided the lowest error and highest PSNR, whereas ResShift produced a slightly higher SSIM (.886 vs .866). All improvements of 7T-Restormer over the baselines are statistically significant (two-tailed Welch's t-tests, p < 0.001 for every metric and field strength). 7T-Restormer reduced NMSE by 64% at 1.5T (0.019 vs. 0.052) and 55% at 3T (0.021 vs. 0.060) relative to the diffusion-based ResShift. Compared to the hybrid CNN–transformer ResViT, NMSE dropped by 41 % at 1.5T and 36 % at 3T. NMSE dropped by 41% at 1.5T (0.019 vs. 0.032) and 36% at 3T (0.021 vs. 0.033). Despite these gains in accuracy, the 7T-Restormer network is substantially more lightweight, with only 10.5 million trainable parameters versus 56.7M for ResViT and 70.4M for ResShift. The efficient design also accelerates inference: our model generates a full 256×384 axial slice in 0.27 s on a single NVIDIA A6000 GPU, which is an order of magnitude faster than ResShift's multi-step diffusion pipeline.

Our proposed method also demonstrated superior visual fidelity in synthesized 7T $T_1$ maps from both 1.5T (Figure 2) and 3T inputs (Figure 3). Qualitatively, all methods produced plausible 7T-like contrast. However, the 7T-Restormer outputs best preserved fine anatomical details. In particular, our model's predictions show crisp cortical–subcortical contrast, sharp delineation of sulcal patterns (Figure 2a), and the most realistic intensity for ventricular CSF (Figure 2b) closely matching the real 7T references. By contrast, the diffusion model, ResShift, tended to overshoot edges – for example, exaggerating gray–white matter boundaries into unnaturally sharp ridges. The ResViT baseline often had the opposite effect, over-smoothing small structures such as deep gray nuclei, thereby losing some contrast. In challenging inferior brain slices near the cerebellum, 7T-Restormer avoids the spurious hallucination artifacts (ring-like distortions) that intermittently appear in ResViT outputs. However, we note that all methods tended to include details from the T1W MRI that did not appear in the 7T T1-map (Figure 2b, posterior region) and would overly smooth or fail to accurately reconstruct complex anatomical structures (Figure 3b, zoomed in region).

3.1 Ablation Studies

An ablation study (Table 2) examined whether joint training on mixed-field data is superior to training on either 1.5T or 3T separately. For 1.5T test slices, restricting training to 1.5 T alone increased NMSE from 0.019 to $0.021 \pm 0.012$ ($p = 1.0 \times 10^{-3}$) and reduced SSIM from 0.861 to $0.856 \pm 0.075$. Training solely on 3T data yielded an identical NMSE ($0.021 \pm 0.012$) and similar PSNR/SSIM, again significantly worse than the mixed-field baseline for NMSE ($p = 1.6 \times 10^{-2}$). For 3T inputs, the 3T-only network matched the mixed model in NMSE ($0.021 \pm 0.016$, $p = .34$) and PSNR, whereas the 1.5T-only network suffered a substantial error increase to $0.029 \pm 0.019$ ($p < .001$). These findings indicate that including both field strengths during training provides the best overall generalization and that knowledge learned from 3T data transfers well to 1.5T inference.

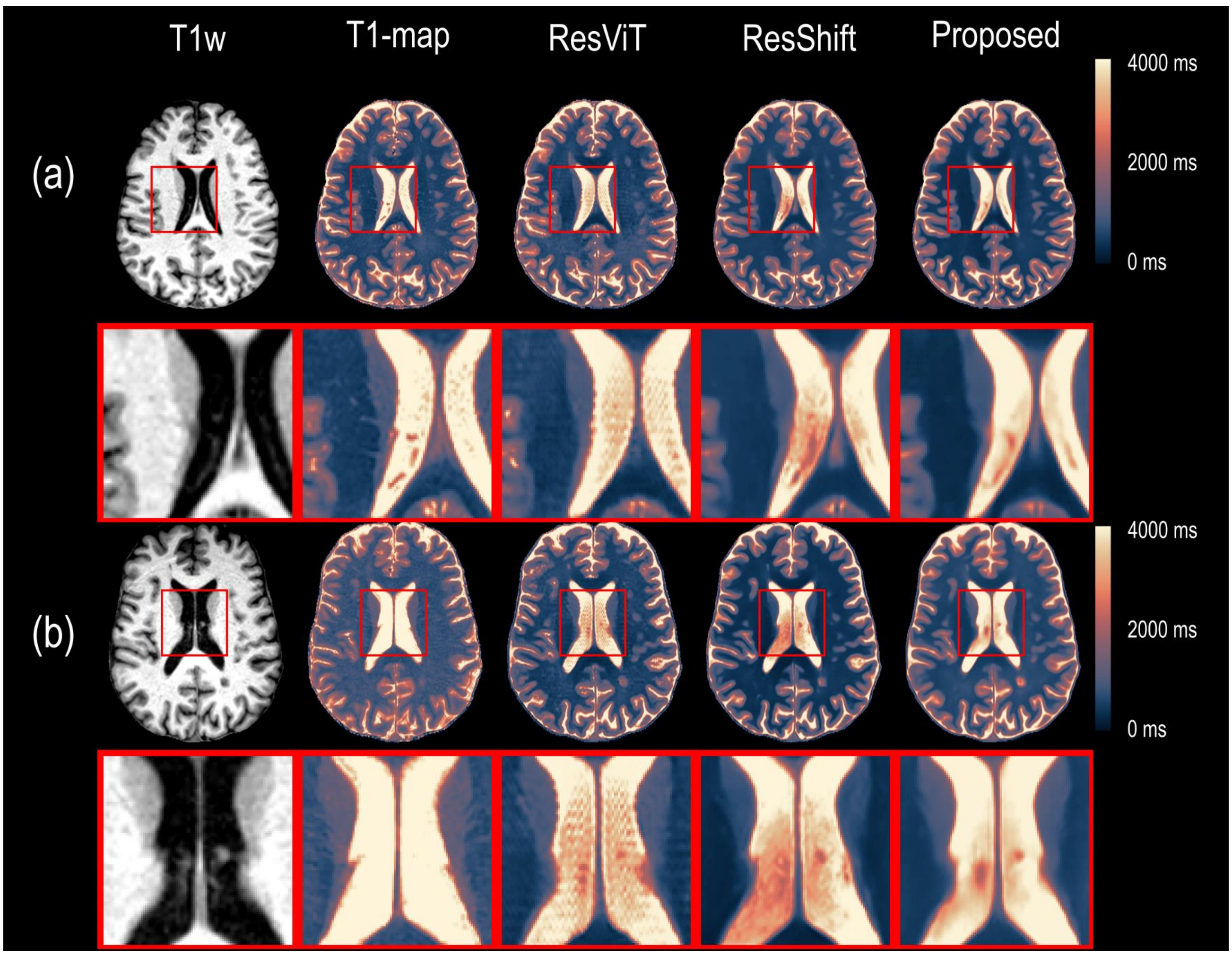

**Figure 2.** Example synthetic 7T T1-maps generated from ResViT, ResShift and the proposed method from 2 patients with 1.5T scans. Zoomed-in regions of interest are featured in the red boxes. Row (a) shows local detail best captured by the proposed method. The box in (b) highlights improved lateral ventricle intensity relative to the ground truth by the proposed model compared to ResViT and ResShift. The scale bar indicates the longitudinal T1 relaxation time in milliseconds.

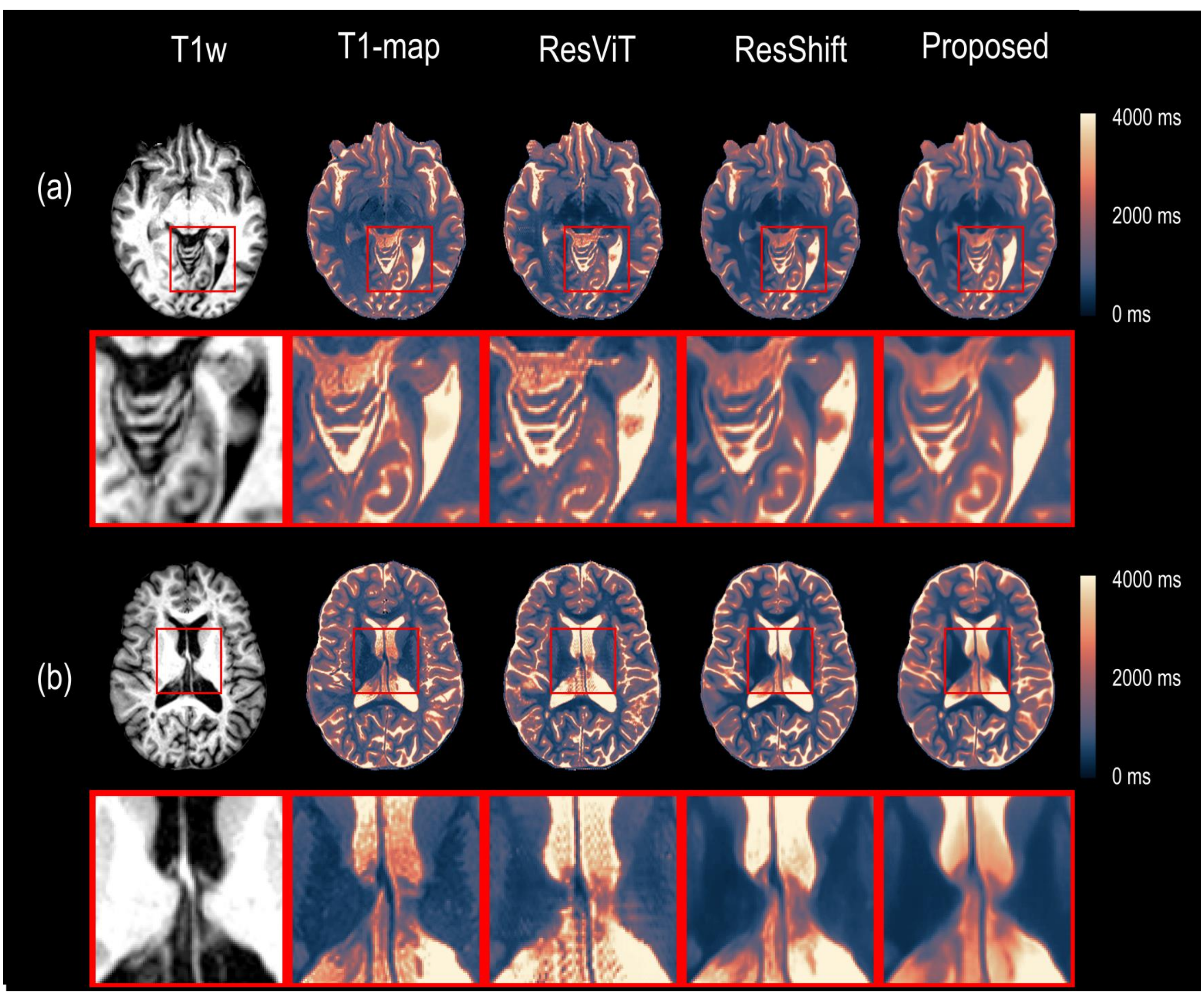

**Figure 3.** Example synthetic 7T T1-maps generated from ResViT, ResShift and the proposed method from 2 patients with 3T scans. Zoomed-in regions of interest are featured in red boxes along with arrows pointing out subtle differences between the methods. Row (a) demonstrates higher conformality to regions around the brain stem. The box in (b) provides an example where the proposed method best captures variations in intensity of the lateral ventricles. The scale bar indicates the longitudinal T1 relaxation time in milliseconds.

**Table 1.** Quantitative results of the proposed method compared to ResShift and ResViT for 1.5T and 3T input modalities. All models performed better for the 1.5T compared to 3T T1W MRI. The proposed model achieved the highest NMSE and PSNR values while ResShift had the highest SSIM values. ResShift-Diff denotes that ResShift is a diffusion model.

| Input | Model | NMSE[↓] | p-value | PSNR (dB) [↑] | p-value | SSIM[↑] | p-value |
|---|---|---|---|---|---|---|---|
| **1.5T** | 7T-Restormer | **0.019 ± 0.011** | x | **26.0 ± 4.6** | x | 0.861 ± 0.072 | x |
| | ResShift-Diff | 0.052 ± 0.010 | <.001 | 25.2 ± 4.7 | <.001 | **0.896 ± 0.052** | <.001 |
| | ResViT | 0.032 ± 0.019 | <.001 | 24.0 ± 4.4 | <.001 | 0.820 ± 0.096 | <.001 |
| **3T** | 7T-Restormer | **0.021 ± 0.015** | x | **25.9 ± 4.9** | x | 0.866 ± 0.077 | x |
| | ResShift-Diff | 0.060 ± 0.025 | <.001 | 24.6 ± 5.0 | <.001 | **0.886 ± 0.062** | <.001 |
| | ResViT | 0.033 ± 0.023 | <.001 | 24.0 ± 4.7 | <.001 | 0.826 ± 0.099 | <.001 |

**Table 2.** An Ablation study showing the effect on different training strategies. Training on both 1.5T and 3T T1W images resulted in greater generalizability than a single input modality alone. 3T only training outperformed 1.5T only for both tasks.

| **Input** | **Training Strategy** | **NMSE[↓]** | **p-value** | **PSNR (dB)[↑]** | **p-value** | **SSIM[↑]** | **p-value** |
|---|---|---|---|---|---|---|---|
| **1.5T** | 1.5T + 3T T1W | **0.019 ± 0.011** | x | **26.0 ± 4.6** | x | **0.861 ± 0.072** | x |
| | 1.5T T1W only | 0.021 ± 0.012 | 0.001 | 25.6 ± 4.6 | 0.086 | 0.856 ± 0.075 | 0.166 |
| | 3T T1W only | 0.021 ± 0.012 | 0.016 | 25.7 ± 4.6 | 0.183 | 0.858 ± 0.076 | 0.386 |
| **3T** | 1.5T + 3T T1W | **0.021 ± 0.015** | x | **25.9± 4.9** | x | **0.866 ± 0.077** | x |
| | 1.5T T1W only | 0.029 ± 0.019 | <.001 | 24.5 ± 5.0 | <.001 | 0.844 ± 0.086 | <.001 |
| | 3T T1W only | 0.021 ± 0.016 | 0.335 | 25.8 ± 4.9 | 0.781 | 0.865 ± 0.080 | 0.485 |

## 4. DISCUSSION

In this study, we propose the 7T-Restormer model for 7T T1-map synthesis which outperforms the state-of-the-art GAN-based ResViT and diffusion ResShift models while avoiding challenges with GANs (training instability and hallucinations) and diffusion models (time consuming inference). By synthesizing 7T-like images from standard (1.5 T) and high field (3T) MRI commonly available in clinical practice, the favorable imaging characteristics of ultra-high field MRI can be made more readily available when 7T MRI scans are impractical to obtain. These high-quality images could be used to enhance diagnostic accuracy, treatment planning, and facilitate downstream tasks.

To our knowledge, this is the first work to synthesize 7T T1-maps from T1W MRI. However, several notable studies have generated ultra-high field MRI from lower field strengths. For example, Qu et al. introduced WATNet, a wavelet-augmented U-Net that increased 3T image resolution and contrast to mimic 7T, achieving a mean PSNR of 28 dB and SSIM of 0.88 on 15 paired brain scans.[8] Expanding upon this framework, Cui et al. integrated a 3D V-Net architecture with adversarial and segmentation losses, further improving the preservation of subtle cortical features across multi-site MRI datasets.[39] A larger clinical study by Duan et al trained a GAN on paired 3T/7T scans and demonstrated that the synthetic 7T T1W images were indistinguishable from true 7T in sharpness, tissue contrast, and artifact level based on blinded radiologist ratings.[40] Several studies have also translated diffusion MRI to higher field strengths. Jha et al. later proposed TrGANet, an ODE-inspired CNN with graph attention that converts single-shell 3T diffusion scans into multi-shell 7T data, recovering fiber orientation distributions nearly indistinguishable from ground truth.[41] Most recently, a hybrid CNN-transformer architecture synthesized 7T apparent diffusion coefficient (ADC) maps from 3T inputs across 171 paired scans, achieving a PSNR of 27 dB and SSIM of 0.95.[12] We note several limitations of these studies. All studies relied extensively on convolutional layers which may struggle with long-range context while several utilized small datasets that tend to hurt generalizability and model performance. In addition, these studies did not investigate the differences between using 1.5T and 3T MRI as model inputs.

Qualitatively, 7T-Restormer preserves fine anatomical details and ventricular CSF intensity more faithfully than the baselines. We identify several factors which provide our model an advantage over baseline methods. First, MDTA blocks calculate self-attention in the channel dimension enabling the model to capture long-range anatomical context while also operating on full-resolution slices. This alleviates the need to first perform the convolution-based downsampling performed in hybrid CNN-transformer architectures. Second, GDFN units re-inject locality, sharpening fine structures that might otherwise be blurred by global attention. Third, our method does not require the stochastic sampling steps intrinsic to diffusion models, significantly decreasing inference time. In addition, both ResViT and ResShift implement generative learning techniques (GAN and diffusion architectures, respectively) which can improve the quality of the images but risk hallucination artifacts. We also note that the ground truth images often contained a non-negligible amount of noise and ResViT's adversarial loss may be injecting this noise at the cost of accuracy as seen in the grainier appear of the zoomed in regions (Figures 2 and 3).

We find that training on mixed 1.5T + 3T data further improved generalization. Restricting optimization to a single field strength consistently raised error on at least one modality, whereas the unified model maintained top performance across domains (Table 2). This suggests that differences in noise statistics and tissue contrast between 1.5T and 3T scans are complementary rather than deleterious when sufficient capacity and augmentation are provided. Additionally, the 3T-only model outperformed 1.5T even on 1.5T MRI. We speculate that this may be due to the larger sample size of 3T (104 3T vs 34 1.5T) and the sharper image quality of the 3T scans. This result may also be helpful for large foundation models when deciding how best to manage images on multiple field strengths.

Several limitations must be acknowledged. The current implementation processes 2D axial slices, thus ignoring through-plane context; extending the architecture to true 3D blocks may further improve anatomical coherence.[42] However, we have found that 3D models effectively reduce the training dataset size from hundreds of slices per patient to, for example, a single 3D volume per patient. For this dataset, we do not believe that we would see a significant improvement and potentially worse performance than a comparable 2D model. In addition, this dataset is comprised of patients with MS, further data from patients with other conditions, such as glioblastoma, will be necessary to achieve similar performance across multiple diseases. All scans were acquired on a single vendor platform, highlighting the need for prospective multi-site validation to confirm robustness across protocol and hardware variability.

Future work will investigate the pretrain and finetune approach common in state-of-the-art foundation models, in which we will first pretrain on a large corpus of brain MR images and subsequently "finetune the model to predict 7T T1 maps[43] This may lead to improved generalizability and compensate for limited data, allowing more effective utilization of 3D models. In addition, we hope to expand our work to generate other 7T sequences and add additional patient data as it becomes available.

## 5. CONCLUSION

This study introduces 7T-Restormer, an efficient transformer-based model that synthesizes high-quality 7T T1-maps from routine 1.5T and 3T T1W MRI. Mixed-field training proved superior to training at different field strengths individually. Our proposed method shows strong potential for making 7T-like scans more readily available and may serve as a foundation for future work in ultra-high field MRI synthesis.

## ACKNOWLEDGMENTS

This research is supported in part by the National Cancer Institute of the National Institutes of Health under Award Numbers R01CA215718, R01EB032680, R21EB033994 and P30 CA008748.

**Disclosures**

The author declares no conflicts of interest.